\documentclass[10pt, a4paper]{article}

\usepackage[]{lrec-coling2024}

\usepackage[utf8]{inputenc}

\usepackage{microtype}
\usepackage{enumitem}
\usepackage{amssymb}
\usepackage{mathrsfs}
\usepackage{amsmath}
\usepackage{hhline}
\usepackage{amsfonts}
\usepackage{multirow}
\usepackage{array}
\usepackage{booktabs}
\usepackage{colortbl}
\usepackage{color}
\usepackage{subcaption}
\usepackage{threeparttable}
\usepackage{float}
\usepackage{caption}
\usepackage{graphicx}
\usepackage{arydshln}
\usepackage{makecell}
\usepackage{inconsolata}
\usepackage{xcolor}
\usepackage{amssymb}
\usepackage{pifont}
\usepackage{xstring}
\usepackage{color}

\newcommand{\cmark}{\ding{51}}%
\newcommand{\xmark}{\ding{55}}%
\definecolor{mygray}{gray}{0.9}
\definecolor{mypink}{rgb}{0.99,0.91,0.95}
\definecolor{mycyan}{cmyk}{0.3,0,0,0}

\usepackage{xcolor}
\usepackage{hyperref}
 \definecolor{darkblue}{rgb}{0, 0, 0.5}
  \hypersetup{colorlinks=true, citecolor=darkblue, linkcolor=darkblue, urlcolor=darkblue}

\title{SciNews: From Scholarly Complexities to Public Narratives \\-- A Dataset for Scientific News Report Generation}

\name{Dongqi Liu, Yifan Wang, Jia Loy, Vera Demberg} 

\address{Department of Computer Science\\
        Department of Language Science and Technology\\
        Saarland Informatics Campus, Saarland University, Germany\\
         \{dongqi, yifwang, jialoy, vera\}@lst.uni-saarland.de\\}
         
\abstract{
Scientific news reports serve as a bridge, adeptly translating complex research articles into reports that resonate with the broader public. The automated generation of such narratives enhances the accessibility of scholarly insights. In this paper, we present a new corpus to facilitate this paradigm development. Our corpus comprises a parallel compilation of academic publications and their corresponding scientific news reports across nine disciplines. To demonstrate the utility and reliability of our dataset, we conduct an extensive analysis, highlighting the divergences in readability and brevity between scientific news narratives and academic manuscripts. We benchmark our dataset employing state-of-the-art text generation models. The evaluation process involves both automatic and human evaluation, which lays the groundwork for future explorations into the automated generation of scientific news reports. The dataset and code related to this work are available at \url{https://dongqi.me/projects/SciNews}.
\\ \newline  }

\begin{document}

\maketitleabstract

\section{Introduction}
\begin{figure}[t]
 \includegraphics[width=1.0\linewidth]{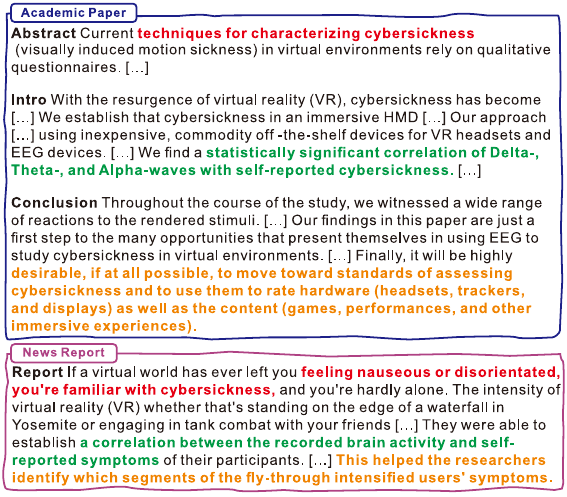}
    \caption{An example of an academic paper paired with its news report.}
    \label{fig:example}
\end{figure}

\textbf{Why Studying Scientific News Report Generation is Valuable:} Scientific publications capture the latest advancements and discoveries in the realm of science, but often necessitate a significant level of academic background, posing obstacles for the general public without specialized knowledge \citep{saikh-etal-2020-scholarlyread, wright-augenstein-2021-semi, august-etal-2022-generating, wright-etal-2022-modeling}. In a bid to bridge this knowledge gap, science journalists are endeavoring to translate intricate scientific nuances and breakthroughs into concise and accessible language \cite{polman2014science, majetic2014science, li-etal-2017-nlp, hoque2022sciev}. This initiative seeks to promote a profound engagement between the public audience and scientific literature \cite{ravenscroft-etal-2018-harrigt, vadapalli-etal-2018-science, august-etal-2020-writing}. Figure \ref{fig:example} illustrates how scientific news reports/narratives may help to increase the accessibility of scientific discoveries by using simplified language, examples, and explanations for technical terms (e.g., ``cybersickness''$\rightarrow$``feeling nauseous or disorientated''). Regrettably, the pursuit of automated generation of scientific news reports faces challenges due to the insufficient availability of parallel corpora. Thus, this paper proposes $(\romannumeral1)$ a new task, Automated Scientific News Report Generation (SNG), and $(\romannumeral2)$ a novel dataset, SciNews, designed for this task.

\textbf{Similarities and Differences with Text Summarization and Text Simplification:} Text summarization emphasizes the reduction of textual volume whilst preserving main information, without altering linguistic complexity \cite{liu-etal-2023-revisiting, pu-etal-2023-incorporating, hosking-etal-2023-attributable, cho-etal-2022-toward, goyal-etal-2022-snac, pu-etal-2022-two, lai-etal-2022-exploration, see-etal-2017-get}, while text simplification focuses on employing simplified lexicon and syntax to enhance readability \cite{pu-demberg-2023-chatgpt, nisioi-etal-2017-exploring, sulem-etal-2018-simple, blinova-etal-2023-simsum, laban-etal-2023-swipe, garimella-etal-2022-text, cripwell-etal-2022-controllable}. The SNG intertwines these processes, requiring both the simplification of complex concepts to more comprehensible forms and the extraction of pivotal insights from source materials \cite{9378403, august-etal-2020-writing, dong-etal-2021-parasci, august-etal-2022-generating, tan-etal-2023-multi2claim}.

Unlike previous efforts mainly focusing on the biomedical field and generating lay summaries of academic paper abstracts \cite{guo2021automated, goldsack-etal-2022-making}, our work across a broader range of scientific disciplines, aims for more comprehensive narrative generation. In addition, the SNG task poses a heightened challenge for text generation models, as it necessitates both a deep understanding of academic discourses and the capability to articulate coherent, long-form articles.

\textbf{Our Contributions:} Given the insufficient availability of benchmark datasets to gauge the potential of text generation models in the SNG task, we present SciNews\footnote{SciNews can only be used for academic purposes.}, a novel multidisciplinary English dataset constructed for automated scientific news report generation. Our dataset leverages academic articles as source inputs, aligned with human-authored scientific news reports as target outputs. Additionally, we conduct extensive evaluations with state-of-the-art (SOTA) Natural Language Generation (NLG) models on SciNews, supplemented with assessments from human evaluators to analyze different perspectives of model outputs. Our findings suggest that the current leading models still struggle with hallucination and factual error problems. Furthermore, compared to human abilities in style-adaptive writing, SOTA NLG models exhibit an inferior capacity for converting complicated texts into understandable narratives. To summarize, our contributions are as follows:

\begin{itemize}[leftmargin=8pt,itemsep=1pt,topsep=1pt,parsep=1pt]
    \item We introduce a task focused on the automated generation of scientific news reports, supported by the SciNews dataset, which contains 41,872 samples.
    \item We undertake both quantitative and qualitative analyses of the SciNews dataset, providing insights into variations in linguistic structure and readability between source articles and target reports.
    \item We evaluate state-of-the-art NLG models on our dataset, finding that the abstractive text generation models surpass the extractive ones on this task.
    \item We offer an error analysis, grounded in human evaluations, identifying primary issues in machine-generated scientific news reports.
\end{itemize}

\section{Related Work}
\label{Related_Work}

In tasks of news-related generation, some investigations have been conducted into the automated generation of general news articles \cite{sigita2013automatic, nesterenko-2016-building, mosallanezhad2020topic, horvitz-etal-2020-context, shu2021fact}, headlines \cite{gusev2019importance, bukhtiyarov2020advances, liu-etal-2020-diverse, ao-etal-2021-pens, panthaplackel-etal-2022-updated, cai-etal-2023-generating}, comments \cite{yang-etal-2019-read}, and summaries \cite{nallapati-etal-2016-abstractive}. Similarly, science-related generation efforts have focused on producing academic summaries \cite{cohan-etal-2018-discourse, cachola-etal-2020-tldr, lu-etal-2020-multi-xscience}, contributions \cite{hayashi-etal-2023-whats}, related work \cite{hu-wan-2014-automatic}, definitions \cite{august-etal-2022-generating}, paraphrases \cite{dong-etal-2021-parasci}, and claims \cite{wright-etal-2022-generating, hayashi-etal-2023-whats, tan-etal-2023-multi2claim}. However, thus far, attempts to study the automated generation of scientific news reports from academic papers across various fields have been less than comprehensive, with a concurrent dataset containing just over 2,400 samples that align with source research papers \cite{cardenas-etal-2023-dont}. In addition, the dataset from \citet{cardenas-etal-2023-dont} primarily focuses on generating press releases from news articles. In contrast, our initiative aims at using academic papers as a foundation to produce news articles. This complementary dataset underscores our shared goal of enhancing public engagement with science, albeit through different lenses of scientific communication. We next explore related areas of scientific lay summarization and text simplification to contextualize our approach within the broader landscape of making science accessible.

\subsection{Scientific Lay Summarization}

Scientific Lay Summarization (SLS) strives to produce accessible summaries that enable researchers in the field to quickly grasp the main content of current papers. For example, at EMNLP 2020, \citet{sdp-2020-scholarly} released a small-scale corpus and organized a shared task. However, this small corpus has proven challenging for training neural NLG models\cite{chandrasekaran-etal-2020-overview}. To alleviate this problem, subsequent studies \cite{guo2021automated, guo2022cells} introduced two larger-scale datasets, demonstrating the effectiveness of neural architectures in SLS. Further contributions \cite{goldsack-etal-2022-making, goldsack-etal-2023-biolaysumm} have added two expansive lay summary datasets focused on bio-medicine, enriching the domain. Additionally, the introduction of RSTformer \cite{pu-etal-2023-incorporating} explored the role of discourse structure in improving SLS. It is essential to note that while current efforts predominantly convert academic paper abstracts into lay summaries, generating scientific news articles requires adopting a narrative-driven approach. This storytelling style poses challenges related to text length and content, typically including aspects such as research background, findings, and impacts.

\subsection{Scientific Text Simplification} 

Scientific Text Simplification (STS) seeks to make complicated texts more readable through text style transfer. Previous attempts, such as the one by \citet{coster-kauchak-2011-simple}, introduced a sentence-level parallel simplification dataset sourced from Wikipedia. Building upon this, \citet{kim-etal-2016-simplescience} established an additional corpus focusing on the lexical simplification of scientific articles, and \citet{grabar-cardon-2018-clear} curated a simplification corpus tailored to Medical French. \citet{laban-etal-2021-keep} devised a reinforcement learning-based system for simplifying multi-sentence structures, while \citet{devaraj-etal-2021-paragraph} applied the Transformer model for paragraph-level simplification of medical texts. Most recently, \citet{ermakova2022automatic, ermakova2023clef} initiated a scientific simplification task at CLEF2022/3. Furthermore, \citet{blinova-etal-2023-simsum} proposed SIMSUM, a strategy for document-level text simplification via simultaneous summarization. However, the majority of the current studies center on simplification at the lexical, sentence, paragraph/short-document levels, leaving substantial unexplored room in long-document simplification \cite{devaraj-etal-2021-paragraph, garimella-etal-2022-text, laban-etal-2023-swipe, cripwell-etal-2023-context, fatima-strube-2023-cross}. In contrast to STS, which simplifies language while preserving academic integrity and depth, ensuring no critical information is lost \cite{cripwell-etal-2023-document}, scientific news narratives, although precise, may alter the depth of discussion and incorporate additional explanatory information for enhanced clarity and reader engagement.

\section{The SciNews Dataset}

\subsection{Task Formulation}

The task of SNG can be formalized as follows: Given a scientific paper \( x_i \) and its corresponding news article \( y_i \), we have a dataset \( D = \{(x_1, y_1), (x_2, y_2), \ldots, (x_n, y_n)\} \), and $(x_i, y_i)$ $\in$ $D$. Our objective is to train NLG models \( N \), such that the model learns a conditional probability distribution \( P(Y|X) \), where \( Y=\{y_1, y_2, \ldots, y_n\} \) and \( X = \{x_1, x_2, \ldots, x_n\} \). When a new \( x_j \) $\notin$ \( X \) is fed into \( N \), it should generate the corresponding \( y_j \).

\subsection{Data Acquisition}

The SciNews is derived from the \href{https://sciencex.com/}{Science X} platform, an important open-access hub featuring news on science, technology, and medical research. It is noteworthy that the news articles on this platform are contributed by authors or their affiliated institutions and are carefully revised by skilled editors to ensure narrative consistency and mitigate potential ethical issues. In compliance with Science X's terms, which permit data collection for academic research without prior written consent, we collect data such as news titles, news content, associated URLs, relevant DOIs, and domain tags for our study.

Our work focuses on the generation of one-to-one news reports, thus we exclude samples derived from multiple research papers. Leveraging the DOI information, we identify and select articles that are open access and published under the ``Creative-Commons'' CC-BY-4.0 license\footnote{\url{https://creativecommons.org/licenses/by/4.0/}}. For data extraction, we employ web scraping tools \href{https://www.selenium.dev/}{Selenium} and \href{https://www.crummy.com/software/BeautifulSoup/}{BeautifulSoup}, which facilitate the retrieval of article content and citation details, ensuring compliance with copyright licenses.

\subsection{Data Cleaning}

We follow the steps of \citet{cohan-etal-2018-discourse, cachola-etal-2020-tldr} to clean the acquired data. In the first phase, we apply PySBD rule-based parser \cite{sadvilkar-neumann-2020-pysbd} and spaCy to remove line breaks, emoticons, and web links from news articles and scientific papers. Next, we delete news reports (and their corresponding papers) associated with multiple disciplines, identified by their domain tags. For academic papers, we limit the extraction to the text between the abstract and the references section. Finally, we exclude documents exceeding 30,000 words, likely dissertations or monographs, and those under 2000 words, typically tutorials or research proposals\footnote{We use \href{https://spacy.io/}{spaCy} to count the number of words.}.

\subsection{Quality Control}
Documents from the Science X platform are high-quality, sourced from reputable academic origins, and authored by both original researchers and professional journalists. Our dataset creation process bypasses the need for further human annotations but incorporates a dual-phase quality control method, including both automated and human assessments. 

\textbf{Automated Quality Control:} Adapting methods from \citet{mao-etal-2022-citesum}, we calculate pairwise BERT similarity score \cite{zhang2019bertscore} between sentences in the news report and their corresponding academic paper. We remove pairs where over half of the news report sentences have BERT similarity scores below 0.5, indicating significant dissimilarity. This procedure is also applied to named entities within the texts, excluding pairs failing to meet this benchmark. Through this vetting process, we remove 612 pairs from our initial set of 42,484 samples.

\textbf{Human Quality Control:} Inspired by \citet{sun-etal-2021-document}, we randomly select 100 article pairs for a manual quality check to evaluate their overall simplicity without sacrificing quality. We utilize a binary judgment to determine if the news narrative is simpler than the academic paper while maintaining its quality. We recruit two evaluators, each having a Master's degree in either Computer Science or Computational Linguistics. Among the 100 samples, only one sample receives divergent assessments -- being labeled as `accepted' by one evaluator and `rejected' by another. The reason given for being `rejected' is that the scientific news report is longer and less concise compared to other test samples, but there are no complaints about other factors, such as simplicity, faithfulness, etc. This sample is retained after a second review confirming its validity. No sample is unanimously rated as `rejected'.

\subsection{Data Splits} 
\begin{figure}[ht]
  \centering \includegraphics[width=0.46\textwidth]{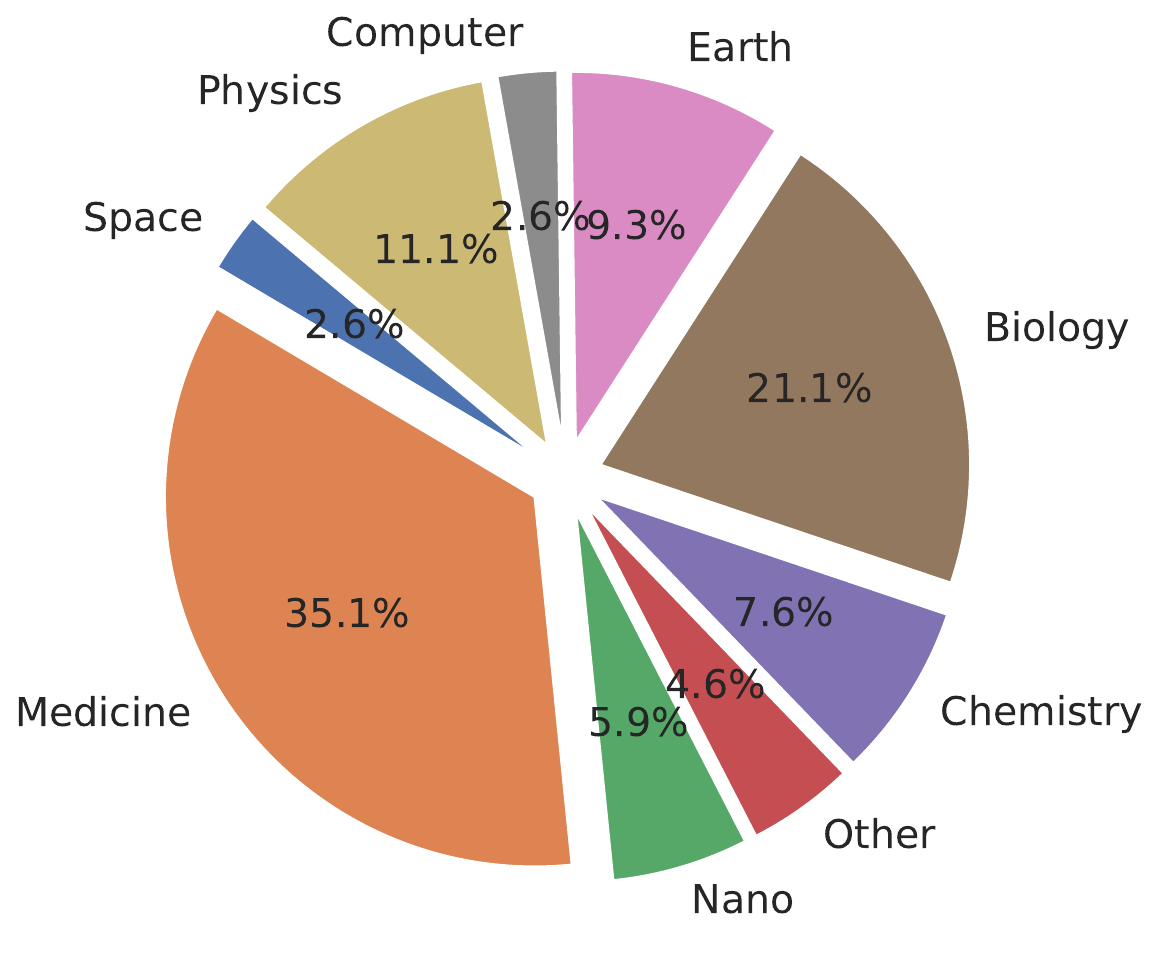}
  \caption{Topic distribution of our dataset}
  \label{fig:topic_distribution}
\end{figure}

After quality control, our dataset comprises 41,872 samples spanning nine scientific domains, as illustrated in Figure \ref{fig:topic_distribution} on topic distribution. We divide the data into training (80\%), validation (10\%), and test set (10\%) by randomly sampling from the entire dataset while keeping the proportion of papers from the different domains constant. The detailed distribution of samples across these subsets is provided in Table \ref{tab:statistics}. All of our experiments described in Sections \ref{sec:experiments} and \ref{sec:results} use this split. 

\begin{table}[h]
\centering
\scalebox{0.96}{
\tabcolsep=4pt
\begin{tabular}{l r}
\toprule
Property & Value \\
\midrule
\# Training Set & 33497 \\
\# Validation Set & 4187 \\
\# Test Set & 4188 \\
\hdashline
Avg. \# Tokens (Papers) & 7760.90 \\
Avg. \# Tokens (News) & 694.80 \\
Avg. \# Sents. (Papers) & 290.52 \\
Avg. \# Sents. (News) & 25.17 \\
\hdashline
Compression Ratio & 12.71 \\
Coverage & 0.74 \\
Density & 0.94 \\
\hdashline
1-gram Novelty & 0.52 \\
2-gram Novelty & 0.91 \\
3-gram Novelty & 0.98 \\
4-gram Novelty & 0.99 \\
\bottomrule
\end{tabular}}
\caption{Dataset statistics}
\label{tab:statistics}
\end{table}

\section{Dataset Analysis}

\subsection{Dataset Comparison}

\begin{table*}[ht]
\centering
\scalebox{0.57}{
\tabcolsep=4pt
\begin{tabular}{l c c c c c c c c }
\toprule
Dataset & Task & Language & Data Scope & Data Source & Scale & Input Level & Output Level & Multi-disciplinary? \\
\midrule
LaySumm \cite{chandrasekaran-etal-2020-overview-insights}  & SLS & English & Archaeology, Hepatology, etc. & Research Papers & 572 & Document & Paragraph & \textcolor{green}{\cmark}\\
CDSR \cite{guo2021automated} & SLS & English & Healthcare & Research Papers & 7805 & Document & Paragraph & \textcolor{red}{\xmark} \\
CELLS \cite{guo2022cells} & SLS &  English & Biomedicine & Research Papers & 47157 & Sentence & Sentence & \textcolor{red}{\xmark} \\
eLife \cite{goldsack-etal-2022-making} & SLS &  English & Biomedicine & Research Papers & 4828 & Document & Paragraph & \textcolor{red}{\xmark} \\
PLOS \cite{goldsack-etal-2022-making} & SLS &  English & Biomedicine & Research Papers & 27525 & Document & Paragraph & \textcolor{red}{\xmark} \\
\hdashline
SimpleScience \cite{kim-etal-2016-simplescience} & STS & English & Biomedicine & Research Papers & 293 & Sentence & Vocabulary & \textcolor{red}{\xmark} \\
CLEAR \cite{grabar-cardon-2018-clear} & STS & French & Biomedicine & Research Papers & 663 & Sentence & Sentence & \textcolor{red}{\xmark} \\
PLS \cite{devaraj-etal-2021-paragraph} & STS & English & Medicine & Research Papers & 4459 & Paragraph & Paragraph & \textcolor{red}{\xmark} \\
SimpleText \cite{ermakova2022automatic, ermakova2023clef} & STS & English & Medicine \& Computer Science & Research Papers & 648 & Sentence & Sentence & \textcolor{green}{\cmark}\\
CSJ \cite{fatima-strube-2023-cross} & STS & English \& German & Astronomy, Biology, etc. & Wikipedia & 50132 & Document & Paragraph & \textcolor{green}{\cmark} \\
\hdashline
SciNews (ours) & SNG &  English & Science \& Technology \& Medicine & Research Papers & 41872 & \cellcolor{mygray}\textbf{Document} & \cellcolor{mygray}\textbf{Document} & \textcolor{green}{\cmark} \\
\bottomrule
\end{tabular}}
\caption{Dataset comparison}
\label{tab:comparison}
\end{table*}

Table \ref{tab:comparison} presents a comparison between our SciNews dataset and datasets for scientific lay summarization and scientific text simplification (as discussed in Section \ref{Related_Work}). Two document-level corpora have a similar size to SciNews (41,872 samples): CSJ has 50,132 samples and PLOS contains 27,525 samples. SciNews stands out due to its multidisciplinary coverage and its provision of category labels for each field. Additionally, the SciNews scientific news reports are longer (averaging 695 tokens), in comparison to PLOS summaries (176 tokens on average), and the simplified texts from CSJ (average length 361 tokens). It is also important to highlight that CSJ derives its data from Wikipedia for multidisciplinary data (without domain labels), in contrast to scholarly articles. Furthermore, CSJ is a paragraph/short-document level simplification dataset, setting it apart from SciNews.

\subsection{Dataset Statistics}
We apply metrics from prior studies \cite{grusky-etal-2018-newsroom, bommasani-cardie-2020-intrinsic, hu-etal-2023-meetingbank} for corpus-level analysis. As Table \ref{tab:statistics} shows, on average, scientific papers consist of 7760.90 tokens and 290.52 sentences, whereas news reports contain an average of 694.80 tokens and 25.17 sentences; the \textit{Compression Ratio} in our dataset is thus 12.71. The \textit{Coverage} metric measures the percentage of tokens in the news report that originate from the original article. A value of 0.74 in \textit{Coverage} indicates substantial inclusion of core information or content from the source in the news articles. The \textit{Density} score assesses the extent to which news reports can be characterized as a set of extractive fragments. The value of 0.94 implies that academic news reports contain only short contiguous text fragments extracted from source papers, indicating a highly abstractive rewriting process.

To measure the textual overlap between news reports and the original papers, we use the methodology from \citet{narayan-etal-2018-dont} and \citet{sharma-etal-2019-bigpatent} to calculate the proportion of 1/2/3/4-grams in news reports that are not present in the original reference texts. The high n-grams novelty scores indicate significant reformation of the material by human authors, suggesting that the news narratives are not just simplified versions of the source texts but involve the creation of novel n-grams through combination, rearrangement, or interpretation of information from the source scientific papers.

\subsection{Papers vs. News}
\label{papersvsnews}

Academic papers typically employ a first-person perspective, in contrast to the third-person narrative found in scientific news articles (as shown in Figure \ref{fig:example}). Beyond the differences in writing tone, we analyze the disparities between these mediums at the lexical (vocabulary), syntactic (sentence)\footnote{\url{https://spacy.io/}}, discourse (intersentential)\footnote{\url{https://github.com/seq-to-mind/DMRST_Parser}} and readability (document)\footnote{\url{https://github.com/textstat/textstat}} levels. As shown in Table \ref{tab:property}, we find that news articles exhibit a higher type-token ratio, indicating greater lexical diversity. Both mediums maintain substantial lexical density, but the news articles contain fewer difficult words.

News articles also use simpler syntactic structures, with fewer modifiers per noun phrase and a reduced average depth of the dependency trees. Moreover, an examination of readability shows a more reader-friendly profile for news texts, corroborated by lower scores in the Flesch-Kincaid Grade Level (FKGL) and the Automated Readability Index (ARI). The statistical significance observed in all metrics of Table \ref{tab:property}, as verified by the Wilcoxon signed-rank test\footnote{\url{https://scipy.org/}} (p$<$0.05), suggests that scientific news narratives function as a more accessible medium with respect to lexical, syntactic and readability features compared to original research papers.

\begin{table}[ht]
\centering
\scalebox{0.9}{\tabcolsep=4pt
\begin{threeparttable}
\begin{tabular}{l r r}
\toprule
Property & Papers  & News \\
\hline
Type-Token Ratio$\uparrow$ & 0.20 & 0.44 \\
Lexical Density$\uparrow$ & 0.42 & 0.46 \\
Avg. \# Difficult Words$\downarrow$ & 773.08 & 134.84 \\
\hdashline
Avg. \# Modifiers per Noun Phrase$\downarrow$ &  0.58 & 0.51 \\
Avg. Depth of Dep Tree$\downarrow$ & 6.94 & 6.25\\
\hdashline
FKGL$\downarrow$ & 14.57 & 13.31\\
ARI$\downarrow$ & 17.94 & 16.32 \\
\bottomrule
\end{tabular}
\end{threeparttable}
}
\caption{Papers and News comparison}
\label{tab:property}
\end{table}

\begin{figure}[t]
  \centering
  \includegraphics[width=0.5\textwidth]{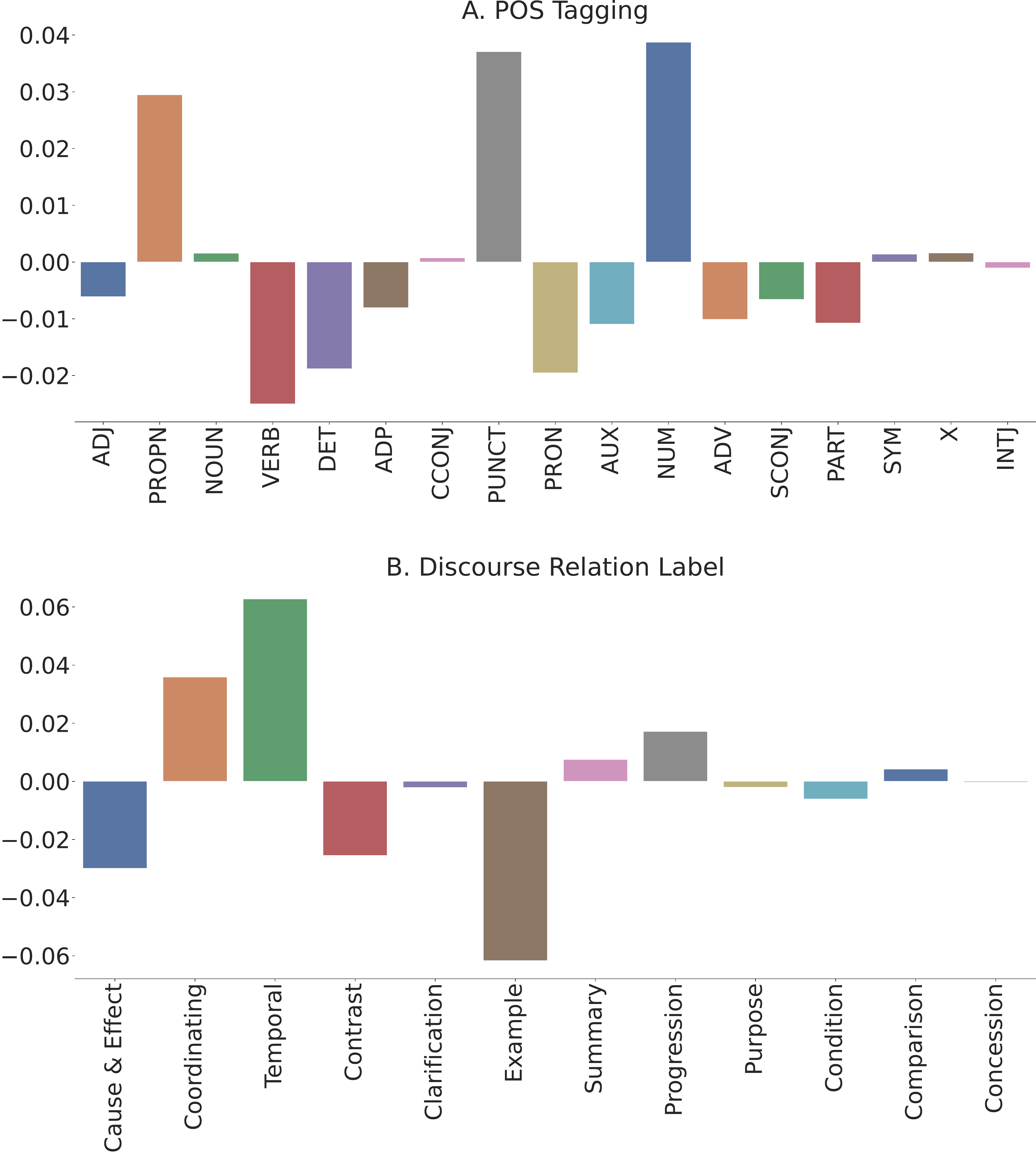}
  \caption{Absolute differences of proportion in linguistic structures (academic papers$-$news articles).}
  \label{fig:difference_list}
\end{figure}

Figure \ref{fig:difference_list}A provides additional details on the distribution of part-of-speech tags between the two text types: news reports contain a higher proportion of verbs and adjectives, while original articles feature more proper nouns, numbers, and punctuation. Regarding rhetorical structure (discourse relations), as shown in Figure \ref{fig:difference_list}B, news reports tend to utilize more `example', `contrast', and `cause \& effect' relations, which may enhance their appeal and accessibility. In contrast, academic texts often favor `temporal', `coordinating', and `progressive' relations to convey research trajectories and findings. 

\section{Experiments}
\label{sec:experiments}
\subsection{Baseline Models}

To promote future work, we benchmark our datasets using two types of baselines: extractive methods and abstractive approaches. Extractive methods involve directly retrieving sentences or phrases from the source text, while abstractive approaches generate outputs by comprehending and paraphrasing the content.

\subsubsection{Extractive Methods}
We select several prevalent algorithms, including heuristic methods like Lead-3/K, Tail-3/K, and Random-3/K (with K denoting the average number of sentences in news reports of the training set, K$=$25), and non-heuristic algorithms like Latent Semantic Analysis (LSA) \cite{steinberger2004using}, LexRank \cite{erkan2004lexrank}, TextRank \cite{mihalcea-tarau-2004-textrank}, Ext-oracle \cite{narayan-etal-2018-dont}, and PacSum \cite{zheng-lapata-2019-sentence}. We also include the original papers as a trivial output to establish the performance of the lower bound for the task.

\subsubsection{Abstractive Methods}
For abstractive methods, we utilize SOTA models based on Seq2Seq like Longformer \cite{Beltagy2020LongformerTL}, RSTformer \cite{pu-etal-2023-incorporating} and SIMSUM \cite{blinova-etal-2023-simsum}, and the Generative Pre-trained Transformer (GPT) architecture like Vicuna7B-16k \cite{zheng2023judging} and GPT-4 \cite{OpenAI2023GPT4TR}. Notably, RSTformer (current SOTA model in SLS task) enhances Longformer's attention mechanism by incorporating discourse knowledge. SIMSUM (current SOTA model in STS task) simplifies documents by aggregating information from several sentences into a single one, omitting some content and breaking down complex sentences. Vicuna7B-16k model is a competitive open-source large language model based on LLaMA~2 \cite{touvron2023llama}. We also conduct a comparison using GPT-4 in a zero-shot setting (ZS).

\subsection{Experimental Settings}
For unbiased comparison, we operate models based on the open-source codes provided by the authors and adhere to the original implementations' default settings, such as model size, batch size, optimizer configuration, and learning rate. During the decoding process, abstractive algorithms are set to a uniform beam search with beam size$=$3 and trigram blocking, we also set temperature and top-$p$ parameters to 1. For Vicuna model, the initial learning rate is set to 5e-5, with a cosine learning rate schedule, batch size of 16, and fully fine-tuned for 30 epochs. The optimizer used is Adam, with \( \beta_1 = 0.9 \), \( \beta_2 = 0.999 \), \( \varepsilon = 10^{-9} \), weight decay = 0.1, and a warm-up ratio of 0.2. To prevent overfitting, we apply early stopping and L2 regularization techniques. Unless stated otherwise, all other parameters align with those in the original publications.

\subsection{Automatic Metrics}

In alignment with other text-generation work \cite{narayan-etal-2018-dont, pu-simaan-2022-passing, liu-etal-2023-binary, pu-etal-2023-incorporating, blinova-etal-2023-simsum}, we examine model performance against the human reference news articles using the following metrics:

\begin{itemize}[leftmargin=8pt,itemsep=1pt,topsep=1pt,parsep=1pt]
    \item ROUGE \cite{lin-2004-rouge} measures the overlap of n-grams between machine-generated output and human-crafted reference. We apply F1 scores of Rouge-1 (R1), Rouge-2 (R2), Rouge-L (RL), and Rouge-Lsum (RLsum) in our analysis.
    \item BERTScore \cite{zhang2019bertscore} examines word overlap between texts, using contextual BERT embedding for semantic similarity analysis.
    \item METEOR \cite{banerjee-lavie-2005-meteor} calculates the harmonic mean of uni-gram precision and recall with an enhanced emphasis on recall for balanced evaluation.
    \item sacreBLEU \cite{post-2018-call} gauges linguistic congruence and translation fluidity between generated and reference texts for comparative analysis of text generation systems.
    \item NIST \cite{lin-hovy-2003-automatic} evaluates the informativeness of n-grams, assigning weights based on corpus frequency-derived information content.
    \item SARI \cite{xu-etal-2016-optimizing} assesses text simplification competency across three dimensions: retention, deletion, and integration of pertinent n-grams for the streamlined rendition of the original text.
\end{itemize}

Additionally, we also use reference-free automatic evaluation metrics from Section \ref{papersvsnews} to evaluate the differences between the top-performing models in their respective categories and human performance on the same test subset. 

\section{Results and Analysis}
\label{sec:results}
\subsection{General Results}

Table \ref{tab:model_performance} depicts the performance of benchmark models on the same test split. Heuristic models such as Lead-3/K, Tail-3/K, and Random-3/K serve as baseline comparison models. Furthermore, we also adopt several popular extractive and abstractive algorithms to explore which algorithm paradigm is more suitable for our dataset. 

\begin{table*}[t]
\centering
\scalebox{0.77}{
\tabcolsep=4pt
\begin{threeparttable}
\begin{tabular}{l c c c c  c  c  c c  c}
\toprule
Model & R1$_{f1}$$\uparrow$ & R2$_{f1}$$\uparrow$ & RL$_{f1}$$\uparrow$ & RLsum$_{f1}$$\uparrow$ & BERTscore$_{f1}$$\uparrow$ & Meteor$\uparrow$ & sacreBLEU$\uparrow$ & NIST$\uparrow$ & SARI$\uparrow$\\

\hline
Full article (lower bound) & 14.42 & 5.21 & 6.90 & 13.94 & 58.55 & 0.21 & 1.49 & 0.55 & 34.83 \\ 
Lead-3 & 14.65 & 4.47 & 8.93 & 13.47 & 54.69 & 0.06 & 0.12 & 0.00 & 35.79 \\ 
Lead-K & 41.99 & 10.96 & 16.13 & 39.68 & 58.55 & 0.27 & 5.25 & 2.34 & 37.21 \\
Tail-3 & 8.43 & 1.46 & 5.41 & 7.77 & 43.61 & 0.03 & 0.05 & 0.01 & 33.94 \\ 
Tail-K & 32.16 & 5.58 & 13.37 & 30.49 & 51.83 & 0.20 & 2.16 & 1.76 & 35.50\\
Random-3 & 10.20 & 1.84 & 6.43 & 9.30 & 47.68 & 0.04 & 0.05 & 0.01 & 34.23\\ 
Random-K & 35.91& 6.90 & 14.10 & 33.83 & 54.41 & 0.22 & 2.68 & 1.97 & 35.94\\
\hdashline
LSA \cite{steinberger2004using} & 39.75 & 8.45 & 15.10 & 37.40 & 56.43 & 0.25 & 3.42 & 2.19 & 36.13\\ 
LexRank \cite{erkan2004lexrank} & 35.59 & 7.98 & 14.97 & 33.62 & 54.49 & 0.24 & 3.22 & 1.92 & 36.16\\ 
TextRank \cite{mihalcea-tarau-2004-textrank} & 35.64 & 7.85 & 14.77 & 33.52 & 53.80 & 0.23 & 3.17 & 1.94 & 36.13 \\ 
PacSum \cite{zheng-lapata-2019-sentence} & 41.03 & 10.53 & 15.47 & 38.75 & 57.64 & 0.27 & 4.82 & 2.28 & 36.85 \\
Ext-oracle \cite{narayan-etal-2018-dont} & 42.58 & 11.92 & 16.16 & 40.38 & 56.60 & \cellcolor{mypink}\textbf{0.30} & 5.90 & 2.43 & 37.28 \\
\hdashline
GPT-4$_{ZS}$ \cite{OpenAI2023GPT4TR} & 41.38 & 9.03 & 15.25 & 39.01 & 58.33 & 0.19 & 4.64 & 1.12 & 37.52 \\ 
SIMSUM \cite{blinova-etal-2023-simsum} & 44.38 & 12.20 & 18.13 & 41.46 & 60.09 & 0.27 & 6.31 & 2.38 & 40.54 \\ 
Longformer \cite{Beltagy2020LongformerTL} & 47.60 & 14.74 & 19.09 & 44.83 & 62.84 & 0.28 & 7.64 & 2.47 & 41.52 \\
RSTformer \cite{pu-etal-2023-incorporating} & \cellcolor{mypink}\textbf{48.21}$^\ddagger$ & \cellcolor{mypink}\textbf{14.92} & \cellcolor{mypink}\textbf{20.12}$^\ddagger$ & \cellcolor{mypink}\textbf{45.19}$^\ddagger$ & 62.80 & 0.28& \cellcolor{mypink}\textbf{7.70} & \cellcolor{mypink}\textbf{2.55} & 41.56 \\
Vicuna7B-16k \cite{zheng2023judging} & 47.75 & 14.88 & 19.92 & 45.01 & \cellcolor{mypink}\textbf{62.88} & \cellcolor{mypink}\textbf{0.30} & 7.69 & 2.53 & \cellcolor{mypink}\textbf{41.71}$^\ddagger$ \\ 
\bottomrule
\end{tabular}
\end{threeparttable}
}
\caption{Model performance. The bold numbers represent the best results with respect to the given test set. $\ddagger$ denotes that the value is significantly superior to those of all other models according to the Wilcoxon signed-rank test in the corresponding indicator (p$<$0.05).}
\label{tab:model_performance}
\end{table*}

Overall, abstractive models significantly outperform both heuristic and extractive models. Specifically, the RSTformer demonstrates superior performance in terms of ROUGE metrics, indicating its enhanced lexical selection capability. Meanwhile, Vicuna surpasses the RSTformer in the SARI metric, highlighting its strengths in simplification and paraphrasing. When it comes to BERTScore, METEOR, sacreBLEU, and NIST metrics, RSTformer and Vicuna exhibit comparable performance.

\subsection{Comparison with Human-authored News Articles}

Table \ref{tab:models_vs_human} contrasts the lexical diversity, syntactic complexity, and readability of the best models for extractive and abstractive methods, as listed in Table \ref{tab:model_performance}, against human ability.

\begin{table}[ht]
\centering
\scalebox{0.68}{
\tabcolsep=4pt
\begin{threeparttable}
\begin{tabular}{lrrrrr}
\toprule
Metric & Human & Ext-oracle  & RSTformer & Vicuna7B\\
\hline
Avg. \# Tokens  & 696.19 & 1274.54 & 653.37 & 782.21\\
Avg. \# Sents.  & 25.29 & 44.51 & 22.85 & 25.03\\
\hline
Type-Token Ratio$\uparrow$ & 0.45 & 0.40 & \cellcolor{mypink}\textbf{0.47} & 0.37 \\
Lexical Density$\uparrow$ & \cellcolor{mypink}\textbf{0.46} & 0.44 & \cellcolor{mypink}\textbf{0.46} & 0.42 \\
Avg. \# Difficult Words$\downarrow$ & \cellcolor{mypink}\textbf{134.65}$^\ddagger$ & 217.37 & 141.75 & 164.5 \\
Avg. \# Modifiers per NP$\downarrow$ & \cellcolor{mypink}\textbf{0.50} & 0.61 & 0.57 & 0.62 \\
Avg. Depth of Dep Tree$\downarrow$ & \cellcolor{mypink}\textbf{6.24}$^\ddagger$ & 6.68 & 7.62 & 6.72 \\
FKGL$\downarrow$ & \cellcolor{mypink}\textbf{13.27}$^\ddagger$ & 15.80 & 14.95 & 14.12 \\
ARI$\downarrow$ & \cellcolor{mypink}\textbf{16.26}$^\ddagger$ & 19.20 & 18.22 & 16.90 \\
\bottomrule
\end{tabular}
\end{threeparttable}
}
\caption{Models vs. Humans; $\ddagger$ indicates that the value significantly differs from those of all other candidates in the same test set, according to the Wilcoxon signed-rank test for the corresponding indicator (p$<$0.05).
}
\label{tab:models_vs_human}
\end{table}

We find that texts generated by the RSTformer model most closely resemble human-written news articles in both length and lexical diversity, while Vicuna-generated texts tend to include slightly longer and more complex words. Additionally, human-written texts are classified as significantly more readable than any model-generated texts, based on FKGL and ARI metrics. Texts generated by Ext-oracle are notable for being much longer and containing more difficult words compared to those written by humans.

\subsection{Automatic Inconsistency Detection}

\begin{figure}[ht]
  \centering
  \includegraphics[width=0.5\textwidth]{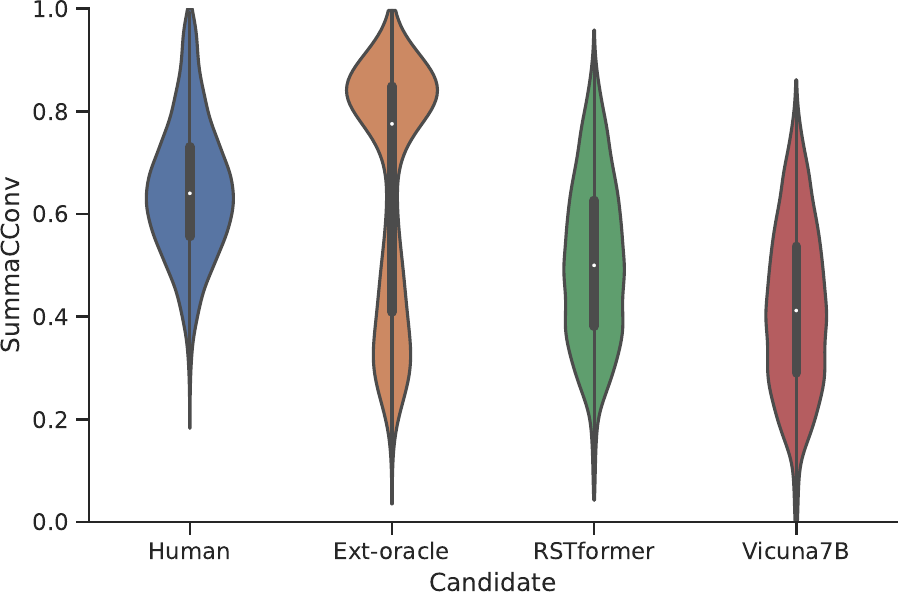}
  \caption{Consistency check}
  \label{fig:summac}
\end{figure}

Figure \ref{fig:summac} shows the outcomes of the automated consistency evaluation for different models on the same test set. We observe that the SummaC consistency scores \cite{laban-etal-2022-summac} for news reports generated by abstractive models fall below those generated by humans in scientific news articles. On the other hand, extractive models, which directly extract text segments from the source, achieve the highest consistency scores without introducing or reorganizing content.

\subsection{Human Evaluation}
\label{sec:human-eval}
In order to gain more insight into the quality of the generated news articles compared to human-authored news articles, we randomly choose 10 samples and present them to human evaluators. The evaluators are asked to read the corresponding original academic article, as well as four candidate news reports (from Ext-oracle, RSTformer, Vicuna, and the original human-authored text). The human evaluators are blind to the condition, i.e., they do not know which article comes from which system (or a human author). Each of the 10 samples is assessed by three different judges, resulting in a total of 30 evaluated samples. The recruited evaluators are all Master’s or Doctoral students with Computer Science or Computational Linguistics backgrounds, with high proficiency in English. All annotators are compensated at the prevailing hourly rates set by the university.

The annotators assess the texts based on the following criteria:
\begin{itemize}[leftmargin=8pt,itemsep=1pt,topsep=1pt,parsep=1pt]
    \item \textbf{Relevant:} How well the news article reflects the source text.
    \item \textbf{Simple:} How understandable the text is for the general public.
    \item \textbf{Concise:} The extent to which the text omits less important information from the source article.
    \item \textbf{Faithful:} The extent to which the text contradicts the information from the source text.
\end{itemize}

Evaluators should assign scores to candidate texts on a scale of 1 to 3 for each criterion, with higher scores indicating better generation quality. Annotators are also required to use different colors to highlight any errors in generated news articles and link them to the corresponding section in the source text. After scoring all candidates, evaluators are asked to identify the best and worst news texts.

Table \ref{tab:Human_evaluation} displays human evaluation results. For each metric, we calculate the average value to assess the candidate system's performance. The terms `Best' and `Worst' denote the frequency with which a model's output is ranked highest or lowest among the four candidates, respectively. Additionally, we count issues flagged by evaluators in the generated news texts, with identical issues highlighted in the same color considered a single instance. 

Ext-oracle performs poorly in terms of `simple' and `concise', as an extractive method, it shows strength in `relevant' and `faithful'. However, it is never chosen as the best candidate by the annotators. RSTformer outperforms Vicuna in `relevant' and `faithful', whereas Vicuna bests RSTformer in `simple' and `concise'. Notably, both abstractive models face challenges with maintaining faithfulness across all generated news texts, a critical issue for practical deployment. Overall, our findings suggest that NLG models have yet to match the proficiency of human writing. This underscores a significant opportunity for future research in enhancing model reliability.

\begin{table}[t]
\centering
\scalebox{0.66}{
\tabcolsep=4pt
\begin{threeparttable}
\begin{tabular}{lccccc}
\toprule
Candidate & Relevant & Simple & Concise & Faithful & Best $\mid$ Worst\\
\hline
Human & \cellcolor{mypink}\textbf{2.67}/$_{0.23}$ & \cellcolor{mypink}\textbf{2.83}$^\ddagger$/$_{0.33}$ & \cellcolor{mypink}\textbf{2.43}$^\ddagger$/$_{0.33}$ & \cellcolor{mypink}\textbf{2.73}$^\ddagger$/$_{0.10}$ & 70.00\% $\mid$ 3.33\% \\
Ext-oracle & 2.63/$_{0.33}$ & 1.30/$_{1.00}$ & 1.20/$_{1.00}$ & 2.63/$_{0.17}$ & 0.00\% $\mid$ 80.00\% \\
RSTformer & 2.63/$_{0.40}$ & 2.27/$_{0.67}$ & 2.03/$_{0.73}$ & 2.17/$_{1.00}$ & 20.00\% $\mid$ 3.33\% \\
Vicuna7B & 2.47/$_{0.60}$ & 2.47/$_{0.67}$ & 2.17/$_{0.60}$ & 1.96/$_{1.00}$ & 10.00\% $\mid$ 13.33\% \\
\bottomrule
\end{tabular}
\end{threeparttable}
}
\caption{Human evaluation results: average ratings (on a scale from 1 to 3). The number following the slash represents the percentage of evaluation samples in which an issue identified by evaluators occurs at least once. 
}
\label{tab:Human_evaluation}
\end{table}

\subsection{GPT-4 Evaluation}

We also employ the same guidelines used for human evaluation to ask GPT-4 via API queries \cite{OpenAI2023GPT4TR} to assess our benchmark models. For consistency, all experiments adhere to OpenAI's default hyper-parameter settings. To ensure no influence from previous interactions, we reset the conversation history before each GPT-4 query. Initially, we sought to confirm whether GPT-4's evaluations align with human judgments on a subset of 10 samples as discussed in Section \ref{sec:human-eval} (maintaining the same ranking of news report scores), and indeed, we find consistent results across all four criteria. Subsequently, we randomly pick an additional 100 samples from the test set, with the results displayed in Table \ref{tab:gpt4_evaluation}.

\begin{table}[t]
\centering
\scalebox{0.7}{
\tabcolsep=4pt
\begin{threeparttable}
\begin{tabular}{lccccc}
\toprule
Candidate & Relevant & Simple & Concise & Faithful & Best $\mid$ Worst\\
\hline
Human & \cellcolor{mypink}\textbf{2.86}$^\ddagger$ & \cellcolor{mypink}\textbf{2.77}$^\ddagger$ & \cellcolor{mypink}\textbf{2.83}$^\ddagger$ & \cellcolor{mypink}\textbf{2.91}$^\ddagger$ & 92.00\% $\mid$ 0.00\% \\
Ext-oracle & 2.73 & 1.73 & 1.55 & 2.70 & 0.00\% $\mid$ 93.00\% \\
RSTformer & 2.69 & 2.41 & 2.42 & 2.47 & 6.00\% $\mid$ 2.00\% \\
Vicuna7B & 2.56 & 2.59 & 2.53 & 2.32 & 2.00\% $\mid$ 5.00\% \\
\bottomrule
\end{tabular}
\end{threeparttable}
}
\caption{GPT-4 evaluation results on 100 samples}
\label{tab:gpt4_evaluation}
\end{table}

According to Table \ref{tab:gpt4_evaluation}, GPT-4's evaluations mirror those of human evaluations. All models under-perform compared to human answers. The scores of GPT-4 for the two SOTA abstractive models are comparable to each other. Across all test samples, GPT-4 prefers the human answer as the best answer, while the extractive method is frequently rated as the worst.

\subsection{Model Errors}
In conjunction with the above-mentioned human evaluation, we conduct a qualitative analysis to identify the prevalent challenges in current models:

\textbf{1. Hallucinations:} Models may produce ungrounded information. For instance, a model might suggest future research areas for chatbots, even if such discussions are absent from the source document.

\textbf{2. Factual Errors:} Models often misstate facts, especially numerical values. For example, in a cancer identification paper, the original mentions sensitivity at 96.7\% and specificity at 97.5\%, but the model reports them as 88.2\% and 98.3\% respectively. 

\textbf{3. Generalization:} While models generally grasp the primary subject, they sometimes diverge into irrelevant specifics. A case in point is a paper on cybersickness, where the model drifts from the main topic into unrelated areas, unlike a focused human-written article.

\section{Conclusion}

We introduce the scientific news report generation task, and present a novel dataset ``SciNews''. This dataset comprises over 40,000 scientific papers spanning nine distinct domains, each paired with a corresponding news report. We conduct an exploratory analysis of the SciNews dataset and provide benchmark results highlighting the challenges faced by current state-of-the-art models. The SciNews dataset not only offers some research prospects, such as fostering the advancement of improved models for scientific news report generation which are faithful to the facts in the original papers but also suggests the potential enhancement of news reports through the integration of relatable explanations. Additionally, beyond its primary purpose, the SciNews dataset can also serve as a valuable resource for other natural language processing tasks, including topic classification and news headline generation.

\section*{Ethical Considerations}
All data in our dataset are sourced from publicly accessible resources, adhering to the respective copyright and web crawling regulations. Each data sample explicitly displays the relevant source URL and author attributions. Moreover, every data sample has undergone rigorous examination and penning by journalists on the Science X website to mitigate ethical or moral apprehensions. Our methodology discerns no privacy infringements during the data processing, experimental analysis, and model training/evaluation phases. Regarding human evaluation, all contributors participate voluntarily and are fairly compensated. We provide a secure and comfortable environment for evaluations, strictly following the ACM Code of Ethics throughout this study's experiments and analyses.

\section*{Limitations}
\textbf{Data:} The SciNews dataset comprises academic papers in English along with their corresponding news reports. Despite the high-quality sourcing of the data, which involves contributions from domain experts, it is possible that biases specific to certain fields persist. Moreover, we only explore scientific news reports in nine research fields, and these data are only a small part of the real-world data and do not contain all of the academic fields, such as Humanities and Social Sciences. The exclusivity of English within our dataset can be perceived as a limitation, as it presently does not incorporate data in other languages.

\textbf{Model:} In our utilization of the SciNews dataset, we have employed several state-of-the-art models, which may carry biases embedded during their pre-training. However, we have not conducted rigorous assessments regarding the magnitude of these biases within the models as it is beyond the scope of this study. Moreover, the data we gathered all originates from online publicly available resources, so we cannot ascertain whether ChatGPT/GPT-4 has been exposed to or trained on our data during their development (data contamination risks). We acknowledge this limitation and earmark this as a potential space for exploration in our future studies. In addition, we also leave the discussion of differences in the performance of NLG models across different disciplines as part of future work.

\textbf{Automated Evaluation:} Despite employing nine popular automated evaluation algorithms systematically assessing the baseline models from various angles on the test set with human gold answers, and contrasting the baseline models with human performance through multiple reference-free metrics, we recognize that all automated metrics have inherent limitations. Consequently, they might not furnish a comprehensive evaluation of the model's performance.

\textbf{Human Assessment:} The size of the data samples used for human evaluation is constrained by the nature of long document generation and the extensive length of the original texts, often spanning multiple pages. Consequently, expanding the evaluation process through means such as crowd-sourcing becomes challenging. As a result, we can only assess a limited set of 10 documents, which may not offer a fully representative view of the entire dataset. While all of our recruited human evaluators are Master's or Ph.D. students, not all of them are domain experts/lay readers, nor can they be experts/lay readers across multiple scientific fields. Therefore, their judgments cannot be solely relied upon.

\section*{Acknowledgements} 
This project has received funding from the European Research Council (ERC) under the European Union’s Horizon 2020 Research and Innovation Programme (Grant Agreement No. 948878). We are grateful to the reviewers and area chairs for their exceptionally detailed and helpful feedback.
\begin{figure}[H] 
\centering
\includegraphics[width=0.75\columnwidth]{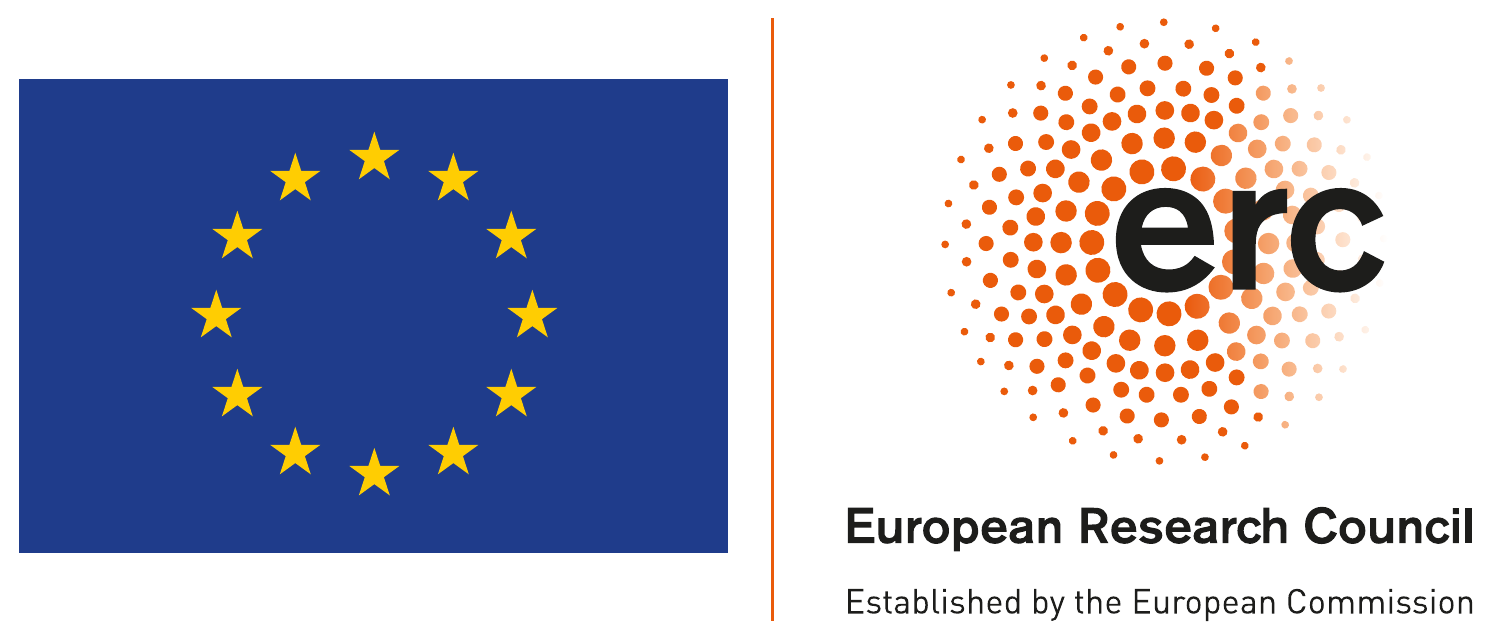}
\end{figure}

\newpage

\nocite{*}
\section{Bibliographical References}
\label{sec:reference}
\bibliographystyle{lrec-coling2024-natbib}
\bibliography{lrec-coling2024}



\end{document}